\pdfoutput=1

\documentclass[11pt]{article}
\usepackage{authblk}
\usepackage{EMNLP2022}

\usepackage{times}
\usepackage{latexsym}
\usepackage{amsfonts}
\usepackage{graphicx}
\usepackage{placeins}
\usepackage{amssymb}
\usepackage{amsmath}
\usepackage{float}
\usepackage[T1]{fontenc}

\usepackage[utf8]{inputenc}

\usepackage{microtype}

\usepackage{inconsolata}

\title{Novel Chapter Abstractive Summarization using Spinal Tree Aware Sub-Sentential Content Selection}

\newcommand\blfootnote[1]{%
  \begingroup
  \renewcommand\thefootnote{}\footnote{#1}%
  \addtocounter{footnote}{-1}%
  \endgroup
}

\author[1]{\textbf{Hardy}\textsuperscript{*}}

\author[2]{\textbf{Miguel Ballesteros}}
\author[3]{\textbf{Faisal Ladhak}\textsuperscript{*}}
\author[4]{\textbf{Muhammad Khalifa}\textsuperscript{*}}
\author[2]{\\\textbf{Vittorio Castelli}}
\author[2]{\textbf{Kathleen McKeown}}
\affil{McGill University, \textsuperscript{2}AWS AI Labs, \textsuperscript{3}Columbia University, \textsuperscript{4}University of Michigan}
\affil[]{\texttt{hardy.hardy@mcgill.ca}, \texttt{ballemig@amazon.com}, \texttt{faisal.ladhak@columbia.edu}}
\affil[]{\texttt{khalifam@umich.edu},
\texttt{vittorca@amazon.com}, \texttt{mckeownk@amazon.com}}

\begin{document}
\maketitle
\begin{abstract}
\blfootnote{\textsuperscript{*}Work done while at Amazon AWS AI Labs}
Summarizing novel chapters is a difficult task due to the input length and the fact that sentences that appear in the desired summaries draw content from multiple places throughout the chapter. We present a pipelined extractive-abstractive approach where the extractive step filters the content that is passed to the abstractive component. Extremely lengthy input also results in a highly skewed dataset towards negative instances for extractive summarization; we thus adopt a margin ranking loss for extraction to encourage separation between positive and negative examples. Our extraction component operates at the constituent level; our approach to this problem enriches the text with spinal tree information which provides syntactic context (in the form of constituents) to the extraction model. We show an improvement of 3.71 Rouge-1 points over best results reported in prior work on an existing novel chapter dataset. 
\end{abstract}

\section{Introduction}

Research on  summarizing novels \citep{Mihalcea2007-jg, WU201712, Ladhak2020-yy, kryscinski2021booksum, wu2021recursively} 
has recently gained popularity following 
advancements in sequence-to-sequence pre-trained models \citep{Zhang2019-mv, Lewis2019-tl, Raffel2019-ym} and in summarization of newswire datasets \citep{Narayan2018-ue, Hermann2015, newsroom_N181065}. Novel chapters present challenges not commonly encountered when summarizing news articles.
Phrases from multiple, non-contiguous sentences within the chapter are often fused to form new sentences for the summary. One would be inclined to use an abstractive approach, but the length of chapters (on average, seven times longer than news articles~\citep{Ladhak2020-yy}) makes it unfeasible to use state of the art generative models, such as BART~\citep{Lewis2019-tl} and even Longformer~\citep{Beltagy2020-sx}. Chapter length causes the additional problem of an imbalanced dataset, as a much higher percentage of the input will not be selected for the summary than is typical in domains such as news. 

To address these challenges, we adopt an extractive-abstractive architecture,  where content is first selected by extracting units from the input and then an abstractive model is used on the filtered  input to produce fluent text.  \citet{kryscinski2021booksum} benchmarked the extractive-abstractive architecture, first proposed by \citet{Chen2018-nt}, for novel summarization, but did not extend it. In this work, we propose several novel extensions to improve its performance on the novel chapter summarization task. 

First, we address the issue of imbalanced dataset where the large amount of compression in novel chapter summarization (372 summary words per 5,165 chapter words on average) creates an extreme imbalance in the training data; a successful extractive summarization algorithm would have to discard most of the text. The standard practice of using Cross-Entropy loss \citep{good1992rational} when training a neural network model backfires in our case: a network that opts to discard everything will achieve near-perfect performance.  We alleviate the issue by improving the margin structure of the minority class boundary using the Margin Ranking loss \citep{rosasco2004loss}, which encourages separation between the two classes. Other studies, such as \citet{cruz2016tackling}, also shows that a pairwise ranking improves model performances on imbalanced data. 

\begin{figure}[!t]
    \centering
    \includegraphics[width=0.46\textwidth]{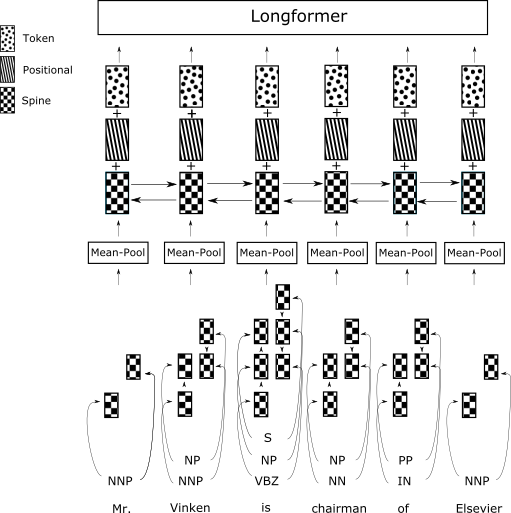}
    \caption{The encoding part of the model for spinal tree encoding. Each token is represented by its corresponding spine from the spinal tree and is encoded by bidirectional-GRU networks \citep{cho2014learning} before being concatenated with its token embedding. We don't show the CLS and SEP tokens here for space-saving purposes, but they are treated as in  BertSumm \citep{liu2019text}.}
    \label{fig:spinaltree}
\end{figure}

Second, in order to model the fusion of chapter phrases into summary sentences, we carry out extraction at the constituent level. \citet{Ladhak2020-yy} also tried this approach, but with mixed results. They noted that sometimes the sub-sentential unit can be too small and, therefore, lack meaningful content (e.g., phrases such as ``what has?'' in the extractive summary, Table~\ref{table:system_examples}). These small unintelligible pieces can negatively affect the performance of the extractive model and, more importantly, the subsequent abstractive model. 
We hypothesize that we can improve the performance of the extractive model---and, consequently, that of the downstream abstractive model---by augmenting the meaning of the extracted sub-sentential units using additional information from the sentence. To that end, we propose an enrichment process, during model training, where we augment the sub-sentential units with linguistic information. For this purpose, we use a spinal tree~\citep{carreras2008tag, Ballesteros2015-wu} which carries information about both the dependency and the constituent structure of the segment. We encode the spine's information using a recurrent network and concatenate its output to the embedding of the token, as illustrated in Figure~\ref{fig:spinaltree}. We choose spinal tree since it 

Our contributions are threefold: (1)  we adopt an extractive-abstractive architecture,   improving the decision boundary of the content selection by using a Margin Ranking loss,  (2) we perform extraction at the constituent level, introducing an enrichment process that uses spinal tree information and (3) we show that our approach improves over the state-of-the-art with a 3.71 gain in Rouge-1 points. 

\begin{table*}[htbp!]
    \centering
    \small
    \begin{tabular}{p{15.5cm}}
        \hline
        \multicolumn{1}{c}{Extracts from our best performance Extractive Model} \\
        \hline
         tess went down the hill to trantridge cross , and inattentively waited to take her seat in the van returning from chaseborough to shaston .<q>her mother had advised her to stay here for the night , at the house of a cottage-woman<q>what has ? ''<q>`` they say -- mrs d'urberville says --<q>that she wants you to look after a little fowl-farm which is her hobby .<q>cried joan to her husband . \\
         \hline
        \multicolumn{1}{c}{Abstracts from our best performance Abstractive Model} \\
        \hline
         tess goes down the hill to trantridge cross and waits to take her seat in the van returning from chaseborough to shaston . her mother has advised her to stay for the night at the house of a cottage-woman who has a fowl-farm . joan tells her husband that mrs. d'urberville has written a letter asking her daughter to look after her poultry-farm  \\
         \hline
         \multicolumn{1}{c}{Reference} \\
         \hline
         when tess returns home the following day. a letter from mrs. d'urberville offering her a job tending fowl awaits her . despite her mother 's ecstatic eagerness , tess is displeased and looks instead for local jobs to earn money to replace the family 's horse .alec d'urberville stops by and prompts her mother for an answer about the job . her efforts to find alternative work prove fruitless and so tess accepts d'urberville 's offer . she remarks that mrs. d'urberville 's handwriting looks masculine .
    \end{tabular}
    \caption{The outputs from two different models. The extract is obtained through a content selection model while the abstract is obtained by passing the extract into BART \citep{Lewis2019-tl} language generation model. The <q> tokens in the extract are the delimiters for constituents.}
    \label{table:system_examples}
\end{table*}

\section{Related Work}
\label{sec:related_works}
Several previous works on novel chapter summarization, such as \citet{Mihalcea2007-jg}, \citet{WU201712}, \citet{Ladhak2020-yy}, \citet{kryscinski2021booksum} and \citet{wu2021recursively}, are closely related to ours. \citet{Mihalcea2007-jg} uses MEAD, an unsupervised extractive summarization described in \citet{radev2004centroid}; this approach includes features focusing on terms weighting that take into account the different topics in the text. In this work, topic boundaries are determined using a graph-based segmentation algorithm that uses normalized cuts \citep{malioutov2006minimum}. A similar line of work, including~\citet{Mihalcea2007-jg} and \citet{WU201712}, also performs topic modelling with Latent Dirichlet Allocation~\citep{blei2003latent} 
followed by greedy unsupervised extraction.

Conversely, \citet{Ladhak2020-yy} 
experiment with extracting information at the sentence and at the syntactic constituent level, via a supervised learning approach. To train their model, they use an aligning process based on the weighted \textsc{ROUGE} scores between the reference and novel text to assign proxy extract labels, in the absence of manually annotated ground truth. Their results at the constituent level are mixed; human evaluation shows a lower performance of constituent extraction models presumably because the summaries are not very readable.  \citet{kryscinski2021booksum} construct a novel chapter dataset that is slightly larger than that of  
 \citet{Ladhak2020-yy} and benchmark existing summarization algorithms on the dataset.

\citet{wu2021recursively}, on the other hand, use a human-in-the-loop approach to obtain summaries via behaviour cloning and reward modelling.

\section{Novel Chapter Summarization}
We use a two-step process  where we first run an \emph{extractive} model \citep{Mihalcea2007-jg,WU201712, Ladhak2020-yy} to select informative content and then run a separate \emph{abstractive} model \citep{Lewis2019-tl, Zhang2019-mv, Raffel2019-ym} to produce a coherent and readable version of this content.

\subsection{Dataset and Pre-processing}
For our novel dataset, we use summary-chapter pairs collected by \citet{Ladhak2020-yy} from Project Gutenberg and various study guide sources. The size of the dataset is 8,088 chapter/summary pairs \footnote{Train/dev/test splits are 6,288/938/862}. The average length of the chapters is 5,165 words with the longest being 33,167 words\footnote{We are aware that there is a larger dataset called BookSum~\citep{kryscinski2021booksum}, which uses similar sources;  however, due to licensing issues, we are unable to use it in our work.}.

In order to prepare the data for the experiments, we follow the same pre-processing steps as \citet{Ladhak2020-yy} to obtain the sub-sentential units and their alignment to reference summaries.  In addition, we truncate chapters to 30k tokens to fit into the GPU memory\footnote{We use Amazon AWS EC2 P4dn 40GB GPU memory}; as a result, a single chapter of the dataset is actually truncated.  

\subsection{Extractive Model}
The extractive summarization task can be posed as a classic regression and ranking problem where the model produces a score for each of a given set of units and then ranks them based on that score. The top $k$ units are then used as an extract. The input of our model is the sub-sentential units of the novel chapter text. We train the model with the oracle labels which we obtain from the alignment between sub-sentential units and reference summaries.  

\paragraph{Baseline}
Our baseline is \textsc{BertSummExt} model~\citep{liu2019text} modified as follows.  First, we replaced the underlying  Transformer models~\citep{vaswani2017attention} with  Longformers, which can better capture long context and requires less computing memory than BERT~\citep{devlin2019bert}.
Second, we removed the inter-sentence Transformer layers stacked on top of the BERT output, to further reduce memory usage. To avoid confusion with \citet{liu2019text}'s model, we named this baseline as Longformer Ext.

\paragraph{Spinal Tree}
A spinal tree is a dependency structure of a sentence that is augmented with constituent information~\citep{carreras2008tag, Ballesteros2015-wu}. For each sub-sentential unit, we retrieve the spinal tree parse by first using the constituency parser \citep{manning-EtAl:2014:P14-5}
and then apply Collins Head-Word Finder~\citep{collins1997three} to calculate the spines. We then encode\footnote{We use the hidden size of 512} the spinal tree using bidirectional-GRU networks~\citep{cho2014learning}\footnote{We experimented with other architectures including bi-LSTM and found that bidirectional-GRU were the best.}.

We construct the input of the Longformer by concatenating the embeddings of the tokens\footnote{We use the embedding size of 768}, the corresponding positional embeddings per token, and the encoding of the spines for each token via the bidirectional-GRU encoders, as illustrated in Figure~\ref{fig:spinaltree}.

\paragraph{Ranking Loss}
The baseline model uses the Cross-Entropy (CE) loss function and minimizes the loss via gradient descent.
However, the CE loss function focuses on optimizing both the negative and positive labels at the same time.  
To compensate for the imbalance in our dataset, we add a Margin Ranking (MR) loss that gives the positive labels higher ranks than the negative labels \footnote{We also have tried the weighted CE loss function but we get worse results. We also found that training our model first with the CE loss function until convergence and then continuing using the MR loss gives the best result.}. 

\paragraph{Re-ordering Scheme}
The default baseline of \citet{liu2019text} produces extracts with sub-sentential units that are ordered based on their score. This scheme, however, destroys the plot of the story. Hence, we re-order the units according to the original positional order in the source text, thus preserving the correct plot order in the story.

\subsection{Abstractive Model}
Since the extractive model outputs are sometimes incoherent and hard to read, we forward them to an abstractive model, with the goal to produce a more fluent and coherent result.

We use BART \citep{Lewis2019-tl} as our engine for abstractive summarization. To train BART, we use the oracle extracts as the input source and the reference summaries as the target. During prediction, we use the output of our content selection model as the input source.
\begin{table}[h!]
\small
\begin{tabular}{p{2cm}p{0.42cm}p{0.42cm}p{0.42cm}p{0.38cm}p{0.4cm}}
\hline
Model                         & \multicolumn{1}{l}{R1} & \multicolumn{1}{l}{R2} & \multicolumn{1}{l}{RL} & \multicolumn{1}{l}{WMD} & \multicolumn{1}{l}{BERTScore} \\
\hline
\multicolumn{6}{c}{Extractive} \\
\hline
Oracle Ext                    & \textbf{46.75}                 & \textbf{14.27}                & \textbf{45.64}                  & \textbf{0.633}                  & \textbf{0.823}                       \\
CB const R-wtd (Ladhak, 2020) & 36.62                  & 6.9                    & 35.4                   & \multicolumn{1}{l}{N/A} & \multicolumn{1}{l}{N/A}       \\
Longformer Ext          & 39.24                  & 7.61                   & 38.29                  & 0.712                  & 0.803                       \\
(Modified \citet{liu2019text})        &                &                   &                 &                 &                      \\
+ Spinal Information           & 39.35                  & 7.62                   & 38.45                  & 0.711                 & 0.802                       \\
+ Ranking Loss                & \textbf{39.48}                  & 7.63                   & \textbf{38.58}                 & \textbf{0.708}                 & 0.802                       \\
+ Re-ordering                      & \textbf{39.48}                 &\textbf{7.70}                  &\textbf{38.58}                 & \textbf{0.708}                 & \textbf{0.806}                     \\
\hline
\multicolumn{6}{c}{Abstractive} \\
\hline
Oracle Abs                    &\textbf{45.82}                 & \textbf{14.14}                 &\textbf{42.74}                 & 0.641                 & 0.828                       \\
BART Abs           & 39.77                  & 9.28                   & 37.56                  & 0.693                 & 0.807                       \\
+ Spinal Information           & 39.83                  & 9.33                   & 37.61                  & 0.691                  & 0.807                       \\
+ Ranking Loss                & 39.88                  & \textbf{9.35}                  & 37.68                  & 0.691                 & 0.807                       \\
+ Re-ordering                      & \textbf{40.33}                 & 9.10                   &\textbf{37.95}                & \textbf{0.690}                 & \textbf{0.810}                   
\end{tabular}
\caption{ROUGE, Word Mover Distance and BERTScore for extractive  and abstractive models}
\label{table:finalresult}
\end{table}

\section{Results}
Examples of outputs from our best abstractive and extractive models are shown in  Table~\ref{table:system_examples}.
Here we report results from an automatic and a manual evaluation.  We compare our approach with and without the different extensions to the prior best model from ~\citet{Ladhak2020-yy}
We also included the oracle for both the extractive and abstractive models. 

\subsection{Automatic Evaluation}
We use three different metrics for automatic evaluation: ROUGE \citep{Lin2004-ek}, BERTScore \citep{zhang2019bertscore} and Word Mover Distance (WMD) \citep{kusner2015word}.  ROUGE measures syntactic similarities between system and reference summaries and BERTScore and WMD measure semantic similarities. BERTScore measures  similarities at the sentence level while WMD does at the token level. We run each experiment three times using  different random seeds and  we report the mean score.

Table~\ref{table:finalresult} shows our models performance against the baseline and previous works. Our best  extractive model (Longformer Ext+spinal+Ranking+Re-ordering) outperforms previous work (CB const R-wtd) by 2.86 ROUGE-1, 0.8 ROUGE-2, and 3.18 ROUGE-L points. Meanwhile, the abstractive model (BART Abs+spinal+Ranking+Re-ordering) outperforms previous work (CB const R-wtd) by 3.71 ROUGE-1, 2.2 ROUGE-2, and 2.55 ROUGE-L points. We have also shown that both the best abstractive and extractive models exceed their corresponding baselines (Longformer Ext and BART Abs) in all metrics. Our models still have room to grow as shown by the oracle results.  

\subsection{Human Evaluation}
For human evaluation, we use the lightweight Pyramid \citep{shapira-etal-2019-crowdsourcing}. We randomly selected 99 samples \footnote{We prepared 100 samples, but one sample got corrupted during the evaluation.} from the test dataset for human evaluation. We also re-run \citet{Ladhak2020-yy}'s output's using the same samples in order to compare ours with their work.

\begin{table}[h!]
\small
\begin{tabular}{ll}
\hline
Model               & Pyramid \\
\hline
CB Const R-wtd (Ladhak, 2020)     & 17.91          \\
BART Abs &  22.03    \\
BART Abs+Spinal+Rank+Re-ordering & 22.86  \\
\end{tabular}
\caption{Pyramid score for our best abstractive performance model compared to previous works}
\label{table:manual}
\end{table}

Table~\ref{table:manual} shows that our models outperform previous work by at least 2 points. We also show that the application of spinal tree enrichment, ranking loss and re-ordering show an improvement of 0.83 points in the human evaluation. 

\section{Conclusion and Future Work}
We have built a novel chapter summarization that produces abstract summaries using a spinal tree aware sub-sentential content selection method. Our results show that we have improved over the state-of-the-art of an existing novel chapter dataset in both automatic and human evaluations.

For future work, we propose an approach where the segmentation of sub-sentential units is jointly trained with the content selection instead of pre-processed before the training process. We hypothesize that this could improve the alignment with reference summaries, therefore, increasing the performance of the overall models.
\section*{Limitation}
The limitation of our work is that the dataset is small. It is also difficult to show significance using a small dataset. Investigation on larger datasets would be necessary to further validate our conclusions.

\section*{Ethical Impact}
We don't foresee any ethical issues with our approach. One could argue that our system might ultimately take jobs away from the people who currently write such summaries. However, given the number of books being written, it is more likely that some summaries would never be written and a good system for novel chapter summarization might help to increase the amount of summaries that are available online. 
\bibliography{custom}

\begin{thebibliography}{31}
\expandafter\ifx\csname natexlab\endcsname\relax\def\natexlab#1{#1}\fi

\bibitem[{Ballesteros and Carreras(2015)}]{Ballesteros2015-wu}
Miguel Ballesteros and Xavier Carreras. 2015.
\newblock Transition-based spinal parsing.
\newblock In \emph{Proceedings of the 19th Conference on Computational Language
  Learning ({CoNLL} 2015). 2015 July 30-31; Beijing, {China.[Stroudsburg]}:
  {ACL}, 2015. p. 289-99.} repositori.upf.edu.

\bibitem[{Beltagy et~al.(2020)Beltagy, Peters, and Cohan}]{Beltagy2020-sx}
Iz~Beltagy, Matthew~E Peters, and Arman Cohan. 2020.
\newblock \href {http://arxiv.org/abs/2004.05150} {Longformer: The
  {Long-Document} transformer}.

\bibitem[{Blei et~al.(2003)Blei, Ng, and Jordan}]{blei2003latent}
David~M Blei, Andrew~Y Ng, and Michael~I Jordan. 2003.
\newblock Latent dirichlet allocation.
\newblock \emph{the Journal of machine Learning research}, 3:993--1022.

\bibitem[{Carreras et~al.(2008)Carreras, Collins, and Koo}]{carreras2008tag}
Xavier Carreras, Michael Collins, and Terry Koo. 2008.
\newblock Tag, dynamic programming, and the perceptron for efficient,
  feature-rich parsing.
\newblock In \emph{CoNLL 2008: Proceedings of the Twelfth Conference on
  Computational Natural Language Learning}, pages 9--16.

\bibitem[{Chen and Bansal(2018)}]{Chen2018-nt}
Yen-Chun Chen and Mohit Bansal. 2018.
\newblock \href {http://arxiv.org/abs/1805.11080} {Fast abstractive
  summarization with {Reinforce-Selected} sentence rewriting}.

\bibitem[{Cho et~al.(2014)Cho, van Merri{\"e}nboer, Gulcehre, Bahdanau,
  Bougares, Schwenk, and Bengio}]{cho2014learning}
Kyunghyun Cho, Bart van Merri{\"e}nboer, Caglar Gulcehre, Dzmitry Bahdanau,
  Fethi Bougares, Holger Schwenk, and Yoshua Bengio. 2014.
\newblock Learning phrase representations using rnn encoder--decoder for
  statistical machine translation.
\newblock In \emph{Proceedings of the 2014 Conference on Empirical Methods in
  Natural Language Processing (EMNLP)}, pages 1724--1734.

\bibitem[{Collins(1997)}]{collins1997three}
Michael Collins. 1997.
\newblock Three generative, lexicalised models for statistical parsing.
\newblock \emph{arXiv preprint cmp-lg/9706022}.

\bibitem[{Cruz et~al.(2016)Cruz, Fernandes, Cardoso, and
  Costa}]{cruz2016tackling}
Ricardo Cruz, Kelwin Fernandes, Jaime~S Cardoso, and Joaquim F~Pinto Costa.
  2016.
\newblock Tackling class imbalance with ranking.
\newblock In \emph{2016 International joint conference on neural networks
  (IJCNN)}, pages 2182--2187. IEEE.

\bibitem[{Devlin et~al.(2019)Devlin, Chang, Lee, and
  Toutanova}]{devlin2019bert}
Jacob Devlin, Ming-Wei Chang, Kenton Lee, and Kristina Toutanova. 2019.
\newblock Bert: Pre-training of deep bidirectional transformers for language
  understanding.
\newblock In \emph{Proceedings of the 2019 Conference of the North American
  Chapter of the Association for Computational Linguistics: Human Language
  Technologies, Volume 1 (Long and Short Papers)}, pages 4171--4186.

\bibitem[{Good(1992)}]{good1992rational}
Irving~John Good. 1992.
\newblock Rational decisions.
\newblock In \emph{Breakthroughs in statistics}, pages 365--377. Springer.

\bibitem[{Grusky et~al.(2018)Grusky, Naaman, and Artzi}]{newsroom_N181065}
Max Grusky, Mor Naaman, and Yoav Artzi. 2018.
\newblock \href {https://doi.org/10.18653/v1/N18-1065} {{Newsroom: A Dataset of
  1.3 Million Summaries with Diverse Extractive Strategies}}.
\newblock In \emph{Proceedings of the 2018 Conference of the North American
  Chapter of the Association for Computational Linguistics: Human Language
  Technologies, Volume 1 (Long Papers)}, volume~1, pages 708--719, Stroudsburg,
  PA, USA. Association for Computational Linguistics.

\bibitem[{Hermann et~al.(2015)Hermann, Ko{\v{c}}isk{\'{y}}, Grefenstette,
  Espeholt, Kay, Suleyman, and Blunsom}]{Hermann2015}
Karl~Moritz Hermann, Tom{\'{a}}{\v{s}} Ko{\v{c}}isk{\'{y}}, Edward
  Grefenstette, Lasse Espeholt, Will Kay, Mustafa Suleyman, and Phil Blunsom.
  2015.
\newblock \href {https://doi.org/10.1109/72.410363} {{Teaching Machines to Read
  and Comprehend}}.
\newblock In \emph{Neural Information Processing Systems}, pages 1--14.

\bibitem[{Kryściński et~al.(2021)Kryściński, Rajani, Agarwal, Xiong, and
  Radev}]{kryscinski2021booksum}
Wojciech Kryściński, Nazneen Rajani, Divyansh Agarwal, Caiming Xiong, and
  Dragomir Radev. 2021.
\newblock \href {http://arxiv.org/abs/2105.08209} {Booksum: A collection of
  datasets for long-form narrative summarization}.

\bibitem[{Kusner et~al.(2015)Kusner, Sun, Kolkin, and
  Weinberger}]{kusner2015word}
Matt Kusner, Yu~Sun, Nicholas Kolkin, and Kilian Weinberger. 2015.
\newblock From word embeddings to document distances.
\newblock In \emph{International conference on machine learning}, pages
  957--966. PMLR.

\bibitem[{Ladhak et~al.(2020)Ladhak, Li, Al-Onaizan, and
  McKeown}]{Ladhak2020-yy}
Faisal Ladhak, Bryan Li, Yaser Al-Onaizan, and Kathleen McKeown. 2020.
\newblock \href {http://arxiv.org/abs/2005.01840} {Exploring content selection
  in summarization of novel chapters}.

\bibitem[{Lewis et~al.(2019)Lewis, Liu, Goyal, Ghazvininejad, Mohamed, Levy,
  Stoyanov, and Zettlemoyer}]{Lewis2019-tl}
Mike Lewis, Yinhan Liu, Naman Goyal, Marjan Ghazvininejad, Abdelrahman Mohamed,
  Omer Levy, Ves Stoyanov, and Luke Zettlemoyer. 2019.
\newblock \href {http://arxiv.org/abs/1910.13461} {{BART}: Denoising
  {Sequence-to-Sequence} pre-training for natural language generation,
  translation, and comprehension}.

\bibitem[{Lin(2004)}]{Lin2004-ek}
Chin-Yew Lin. 2004.
\newblock {{ROUGE}}: A package for automatic evaluation of summaries.
\newblock In \emph{Text Summarization Branches Out}, pages 74--81, Barcelona,
  Spain. Association for Computational Linguistics.

\bibitem[{Liu and Lapata(2019)}]{liu2019text}
Yang Liu and Mirella Lapata. 2019.
\newblock Text summarization with pretrained encoders.
\newblock In \emph{Proceedings of the 2019 Conference on Empirical Methods in
  Natural Language Processing and the 9th International Joint Conference on
  Natural Language Processing (EMNLP-IJCNLP)}, pages 3730--3740.

\bibitem[{Malioutov(2006)}]{malioutov2006minimum}
Igor Igor~Mikhailovich Malioutov. 2006.
\newblock \emph{Minimum cut model for spoken lecture segmentation}.
\newblock Ph.D. thesis, Massachusetts Institute of Technology.

\bibitem[{Manning et~al.(2014)Manning, Surdeanu, Bauer, Finkel, Bethard, and
  McClosky}]{manning-EtAl:2014:P14-5}
Christopher~D. Manning, Mihai Surdeanu, John Bauer, Jenny Finkel, Steven~J.
  Bethard, and David McClosky. 2014.
\newblock \href {http://www.aclweb.org/anthology/P/P14/P14-5010} {The
  {Stanford} {CoreNLP} natural language processing toolkit}.
\newblock In \emph{Association for Computational Linguistics (ACL) System
  Demonstrations}, pages 55--60.

\bibitem[{Mihalcea and Ceylan(2007)}]{Mihalcea2007-jg}
Rada Mihalcea and Hakan Ceylan. 2007.
\newblock Explorations in automatic book summarization.
\newblock In \emph{Proceedings of the 2007 Joint Conference on Empirical
  Methods in Natural Language Processing and Computational Natural Language
  Learning ({{EMNLP}-{C}o{NLL}})}, pages 380--389, Prague, Czech Republic.
  Association for Computational Linguistics.

\bibitem[{Narayan et~al.(2018)Narayan, Cohen, and Lapata}]{Narayan2018-ue}
Shashi Narayan, Shay~B Cohen, and Mirella Lapata. 2018.
\newblock \href {http://arxiv.org/abs/1808.08745} {Don't give me the details,
  just the summary! {Topic-Aware} convolutional neural networks for extreme
  summarization}.

\bibitem[{Radev et~al.(2004)Radev, Jing, Sty{\'s}, and Tam}]{radev2004centroid}
Dragomir~R Radev, Hongyan Jing, Ma{\l}gorzata Sty{\'s}, and Daniel Tam. 2004.
\newblock Centroid-based summarization of multiple documents.
\newblock \emph{Information Processing \& Management}, 40(6):919--938.

\bibitem[{Raffel et~al.(2019)Raffel, Shazeer, Roberts, Lee, Narang, Matena,
  Zhou, Li, and Liu}]{Raffel2019-ym}
Colin Raffel, Noam Shazeer, Adam Roberts, Katherine Lee, Sharan Narang, Michael
  Matena, Yanqi Zhou, Wei Li, and Peter~J Liu. 2019.
\newblock \href {http://arxiv.org/abs/1910.10683} {Exploring the limits of
  transfer learning with a unified {Text-to-Text} transformer}.

\bibitem[{Rosasco et~al.(2004)Rosasco, De~Vito, Caponnetto, Piana, and
  Verri}]{rosasco2004loss}
Lorenzo Rosasco, Ernesto De~Vito, Andrea Caponnetto, Michele Piana, and
  Alessandro Verri. 2004.
\newblock Are loss functions all the same?
\newblock \emph{Neural computation}, 16(5):1063--1076.

\bibitem[{Shapira et~al.(2019)Shapira, Gabay, Gao, Ronen, Pasunuru, Bansal,
  Amsterdamer, and Dagan}]{shapira-etal-2019-crowdsourcing}
Ori Shapira, David Gabay, Yang Gao, Hadar Ronen, Ramakanth Pasunuru, Mohit
  Bansal, Yael Amsterdamer, and Ido Dagan. 2019.
\newblock \href {https://doi.org/10.18653/v1/N19-1072} {Crowdsourcing
  lightweight pyramids for manual summary evaluation}.
\newblock In \emph{Proceedings of the 2019 Conference of the North {A}merican
  Chapter of the Association for Computational Linguistics: Human Language
  Technologies, Volume 1 (Long and Short Papers)}, pages 682--687, Minneapolis,
  Minnesota. Association for Computational Linguistics.

\bibitem[{Vaswani et~al.(2017)Vaswani, Shazeer, Parmar, Uszkoreit, Jones,
  Gomez, Kaiser, and Polosukhin}]{vaswani2017attention}
Ashish Vaswani, Noam Shazeer, Niki Parmar, Jakob Uszkoreit, Llion Jones,
  Aidan~N Gomez, {\L}ukasz Kaiser, and Illia Polosukhin. 2017.
\newblock Attention is all you need.
\newblock In \emph{Advances in neural information processing systems}, pages
  5998--6008.

\bibitem[{Wu et~al.(2021)Wu, Ouyang, Ziegler, Stiennon, Lowe, Leike, and
  Christiano}]{wu2021recursively}
Jeff Wu, Long Ouyang, Daniel~M Ziegler, Nisan Stiennon, Ryan Lowe, Jan Leike,
  and Paul Christiano. 2021.
\newblock Recursively summarizing books with human feedback.
\newblock \emph{arXiv preprint arXiv:2109.10862}.

\bibitem[{Wu et~al.(2017)Wu, Lei, Li, Huang, Zheng, Chen, and Xu}]{WU201712}
Zongda Wu, Li~Lei, Guiling Li, Hui Huang, Chengren Zheng, Enhong Chen, and
  Guandong Xu. 2017.
\newblock \href {https://doi.org/https://doi.org/10.1016/j.eswa.2017.04.054} {A
  topic modeling based approach to novel document automatic summarization}.
\newblock \emph{Expert Systems with Applications}, 84:12--23.

\bibitem[{Zhang et~al.(2019{\natexlab{a}})Zhang, Zhao, Saleh, and
  Liu}]{Zhang2019-mv}
Jingqing Zhang, Yao Zhao, Mohammad Saleh, and Peter~J Liu. 2019{\natexlab{a}}.
\newblock \href {http://arxiv.org/abs/1912.08777} {{PEGASUS}: Pre-training with
  extracted gap-sentences for abstractive summarization}.

\bibitem[{Zhang et~al.(2019{\natexlab{b}})Zhang, Kishore, Wu, Weinberger, and
  Artzi}]{zhang2019bertscore}
Tianyi Zhang, Varsha Kishore, Felix Wu, Kilian~Q Weinberger, and Yoav Artzi.
  2019{\natexlab{b}}.
\newblock Bertscore: Evaluating text generation with bert.
\newblock \emph{arXiv preprint arXiv:1904.09675}.

\end{thebibliography}
\bibliographystyle{acl_natbib}

\appendix
\label{sec:appendix}
\section{Human Evaluation}
For performing Human Evaluation, we use crowdsourcing service provided by Appen.\footnote{\url{https://appen.com/solutions/crowd-management/}} We follow \citet{Ladhak2020-yy}'s approach including the instructions and number of crowdworkers. For each crowdworker, we calculate the payment based on the minimum wage in the US (\$ 15 per hour).

\end{document}